\documentclass{aip-cp}

\usepackage{etoolbox}
\makeatletter
\patchcmd{\@@tablenote}{\xdef}{\protected@xdef}{}{}
\makeatother
\usepackage[font=small,labelfont=bf,justification=centering]{caption}
\usepackage[numbers,elide]{natbib}
\usepackage{fancybox}
\usepackage{shortvrb}
\MakeShortVerb\|
\usepackage{subfig}
\usepackage{listings}
\usepackage[usenames]{color}
\definecolor{codecolor}{gray}{.9}
\definecolor{rlcolor}{cmyk}{0,1,0,0}
\begin{document}

\title{Comparison of Different Control Theories on a Two Wheeled
Self Balancing Robot}

\author[aff1]{MD Muhaimin Rahman}
\eaddress{sezan92@gmail.com}
\author[aff1]{SM Hasanur Rashid}
\eaddress{hrshovon@gmail.com}
\author[aff2]{KM Rafidh Hassan}
\eaddress{rafidhhassan@gmail.com}
\author[aff3]{M. M. Hossain}
\eaddress{monir.eee.buet@gmail.com}
\affil[aff1]{Department of Mechanical Engineering, Bangladesh University of Engineering and Technology}
\affil[aff2]{Department of Mechanical Engineering, Auburn University}

\affil[aff3]{Department of Electrical \& Electronic Engineering, Bangladesh University of Engineering and Technology}

\maketitle

\begin{abstract}
This paper is aimed to discuss and compare three of the most famous  Control Theories on  a Two wheeled Self Balancing Robot Simulation using Robot Operating System (ROS) and Gazebo. Two
Wheeled Self Balancing Robots are one of the most fascinating applications of Inverted Pendulum System. In this paper, PID, LQR and Fuzzy logic controllers are  discussed . Also,0 the modeling and algorithms of the robot simulation is discussed. The primary objectives of this paper is to discuss about the building of a  robot model in ROS and Gazebo , experimenting different control theories on them, documenting the whole process with the analysis of the robot and comparison of different control theories on the system. 
\end{abstract}

\section{INTRODUCTION}
Inverted Pendulum is one of the most discussed and fascinating topics of Non-linear control System in the academia. Many researches have been done on this problem. According to Boubaker et al, no less than ten types of Control theories have been researched on , explored using this problem \cite{man:InvertedPendulum}. One of the most interesting applications of this theory is the Two Wheeled Self balancing Robot. Many researches have been done on this robot also. Like, \cite{man:Hau-Shiue} discussed about developing simple Self balancing robot from popular micro controller board, Arduino. In \cite{man:Xian-gang}, a "Hopfield Neural Network" based controller. In \cite{man:Jianhai} , same robot was developed using Sensor Fusion Algorithm. Again , in \cite{man:Jian} , LQR based controller was used for this robot. In \cite{man:ACM} , self balancing robot was developed using Boltzmann machine. This project is different than others in main purpose. The purpose of this paper is to describe the development of ready to use control algorithms using ROS and Gazebo simulation. ROS and Gazebo are state-of-the-art Open Source Robotics Simulation platform. But there is hardly any good resource on the ROS and control theory , let alone complex non-linear controllers. So , this research opened a door for ROS and Control Systems future Robotics researchers.
\\
The paper is structured as following , first section discusses ROS and Gazebo, second section discusses Experimenting Control system packages with ROS and Gazebo. They are PID, Fuzzy Logic and LQR controllers. The fourth section sheds light on comparison to the available works and literature on same topic and the final section is the conclusion.

\begin{table}[t]
\centering
\captionsetup{justification=centering}
\caption{Nomenclature}
\label{tab:1}
\begin{tabular}{llll}

\hline
M    & Mass of the cart, $0.0754 kg$                                     & I       & Mass moment of inertia 0.01094 $kgm^2$\\ 
m    & Mass of the pendulum                                 & b       & Dynamic Friction Coefficient  = 0.65                       \\ 
l    & Length of the pendulum , 0.157 $m$                              & g       & Gavitational Acceleration= 9.8 $ms^-2$                     \\ 
$P(s)$ & Transfer Function of the system                      & $\phi(s)$ & Transfer function of the yaw angle 
\\ 
$U(s)$ & Transfer Function of the Force  &         &                                                            \\ 
\end{tabular}
\end{table}


\section{ROBOT OPERATING SYSTEM AND GAZEBO}

Robot Operating System (ROS) \cite{man:ROS} is arguably the most powerful tool for modern Robotics Software development. It is a collection of Open Source frameworks, tools ,libraries for research and development of modern robots . Like Universal Robots, PR2 etc are using ROS at present. In this project we are using ROS and Gazebo simulation. Because of following reasons, 
\begin{itemize}
\item It is Open Source, hence it has a vast working frameworks available.
\item We want Robotics Community use and explore various control theories. 
\item But, there is hardly any good resources on ROS Control Theory publicly available.

\end{itemize}

For these reasons, we have developed the self balancing robot Gazebo \cite{man:Gazebo} model. Gazebo is also arguably the best Robotics Simulator at present. Its widely used with ROS interface. It is also open source, which makes
it widely accessible. Figure~\ref{fig:GazeboModel}(b) shows the Gazebo Model of The Robot.
\\
\subsection{Building the Robot Model}
To build the robot Model, the required steps are,
\begin{itemize}
\item Defining Links
\item Defining Joints
\item Adding ROS Plugins
\item Writing the ROS code to test the model
\end{itemize}
The detailed description is given below
\subsubsection{Link}
Link is the building block of every Gazebo simulation . Every robot model will have several types of links. Links can be defined in two ways.
\begin{itemize}
\item Writing xml code from beginning
\item Adding a CAD model in mesh file format. 
\item Combination of Both
\end{itemize}

For building the model by writing xml code , we need to define every aspect of links. Like Mass, geometry, moment of Inertia around every axis etc. It becomes almost impossible for complex links. In that case, the best way is to integrate CAD files of the links.
Another way is to integrate the CAD files as well as defining the simpler links in the xml code.\cite{man:GazeboRef}

In our model of robot, we have three links, Two wheels and a chassis.
A sample xml code for the Right wheel of the robot is given below,
\begin{verbatim}
	<link name = "Right_Wheel">
		<inertial>
			<origin xyz ="0.00091 0.01535 0.07450" rpy = "0 0 0"/>
			<mass value ="0.03770"/>
			<inertia ixx ="0.000233" ixz = "0.000002560" iyy = "0.000236407" iyz = "0.000043120" 
 ixy = "0.000000528" izz = "0.000024435"/>
		</inertial>
		<collision>
			<geometry>
			<cylinder radius="0.0316" length ="0.037"/>	
			
			</geometry>
			<!--
			<origin rpy = "0 0 0" xyz = "0 0.1100 0.0316"/> -->
			<origin rpy = "0 0 0" xyz = ".005 0.160 0.010"/> 

		</collision>
		<visual>
			<geometry>
			<mesh filename="package://self_balancing_robot/meshes/WheelZTiny.dae"/>
			</geometry>
			
			<origin rpy = "0 0 0" xyz = ".005 0.160 0.010"/>  
			
		</visual>
	</link>
\end{verbatim}

\subsubsection{Joint}

\begin{figure}
\label{fig:GazeboModel}
\begin{tabular}[b]{c}
\includegraphics[width=0.5\textwidth]{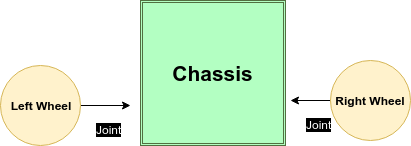}\\
\small (a) Block Diagram for Gazebo Model
\end{tabular}
\begin{tabular}[b]{c}
\includegraphics[width=0.3\textwidth]{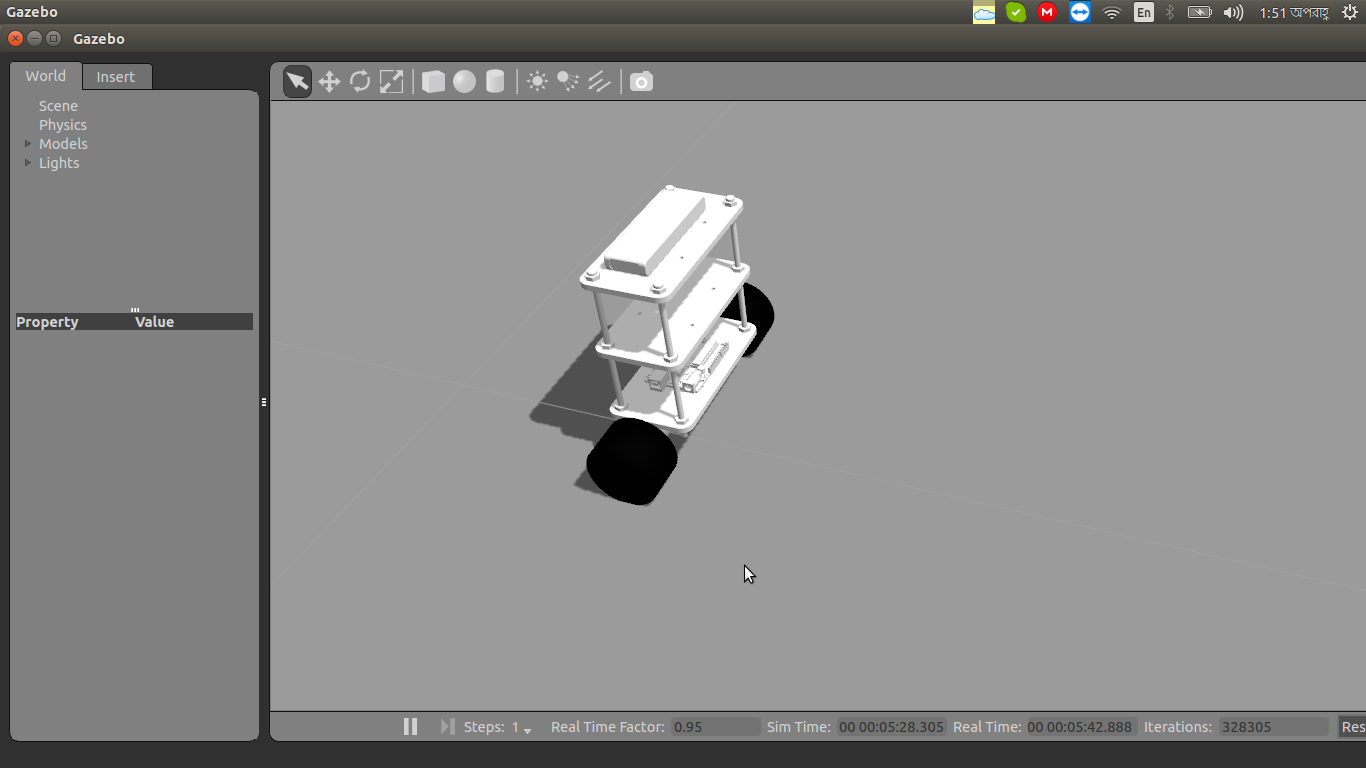}\\
\small (b) Gazebo Simulation
\end{tabular}\\
\caption{Gazebo Simulation Block Diagram and Model}

 \end{figure}

Joint tags define how the links are attached to the base link and other links. \cite{man:GazeboRef} According to \cite{man:ROSJoint}, There are 6 types of joints
\begin{itemize}
\item \emph{Revolute} - links can rotate around specified axis within limits.
\item \emph{Continuous} - It is like revolute type of joint but it has no limits
\item \emph{Prismatic} - Prismatic joint defines the joints where one link can slide on other
\item \emph{Fixed} - Fixed joints are defined for links which cannot move with respect to other
\item \emph{Floating} - Floating type of joints allow movement around six axis of rotation. 
\item \emph{Planar} - Planar type of joints allow motion perpendicular to a plane .
\end{itemize}
In our robot, the wheels are attached to the main body using continuous joint. The figure ~\ref{fig:GazeboModel} Shows the Block Diagram and the Gazebo model.

\begin{figure}
\includegraphics[width=\textwidth]{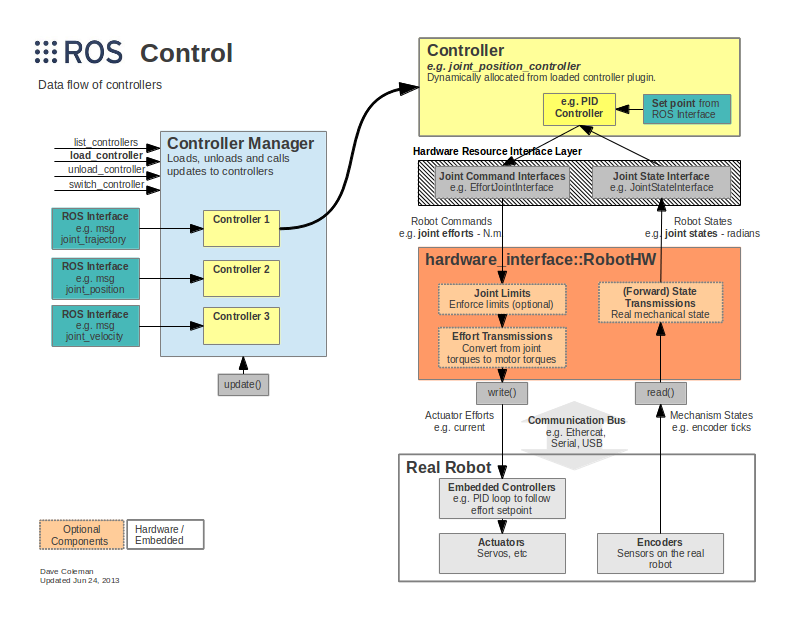}\\
\caption{ROS Controller Application Block Diagram}
\label{fig:ROSControl}
 \end{figure}

 \subsubsection{ROS Plugins}
Plugins make the robot models function according to ROS commands.Without plugins we cannot operate the actuators of the model and we cannot access the sensor outputs.
There are many built in plugins for ROS and Gazebo.\cite{man:GazeboPlugin} We have used Differential Drive Plugin to provide differential drive movement for the robot and Imu Plugin for Imu sensor of the robot.

\section{EXPERIMENT WITH ROS AND OUR CONTROL SYSTEM PACKAGE}
In this section the implementation and performance of three controllers , namely PID, Fuzzy Logic and LQR on the Robot simulation have been discussed using ROS and third party Control System Packages like \cite{man:PIDLibrary}, \cite{man:PythonControlLQR} and \cite{man:scikit-fuzzy}. 
The general algorithm for implementing Controllers is given below
\begin{itemize}
\item Measure the Pitch Angle of the Robot and save it to value y
\item Set Previous Error to Zero
\item Set Sum of Errors to Zero
\item Current Error = Pitch value-Setpoint
\item Error Difference = Current Error - Previous Error
\item Sum of Errors = Current Error+Sum of Errors
\item Get Output from controllers for different Parameters
\end{itemize}
\subsection{PID}
\begin{figure}
\begin{tabular}{c}
\includegraphics[width=0.5\textwidth]{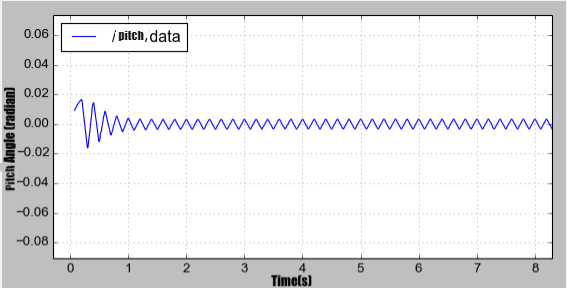}\\
\small (c) Kp=100 Ki=0.8 Kd =0.1
\end{tabular}
\begin{tabular}{c}
\includegraphics[width=0.5\textwidth]{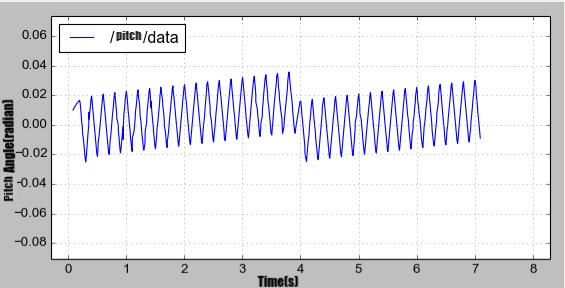}\\
\small (d) Kp=1000, Ki =0.8, Kd =0.05
\end{tabular}
\caption{PID Response Curves for Four different values}
\label{fig:PID}
\end{figure}

Proportional Integral and Derivative (PID) is the most widely used Controller in the world. Li Yun et al discussed about this controller extensively in their paper \cite{man:PID}. In our previous paper \cite{man:SelfBalance} , we have discussed about applying PID controllers in a real Self Balancing Robot using Simulink. In this paper, we are showing the performance of PID packages \cite{man:PIDLibrary} . It follows the equation ~\ref{eq:PID} . Figure ~\ref{fig:PID} show the respose curves for various PID gains. The curves are plotted the $rqt\_plot$ , which also can be used for Real Robot's response plotting.
\begin{equation}\label{eq:PID}
U = K_p*e+K_i\int edt + K_d\frac{de}{dt}
\end{equation}

\begin{figure}

\begin{tabular}{c}
\includegraphics[width=0.5\textwidth]{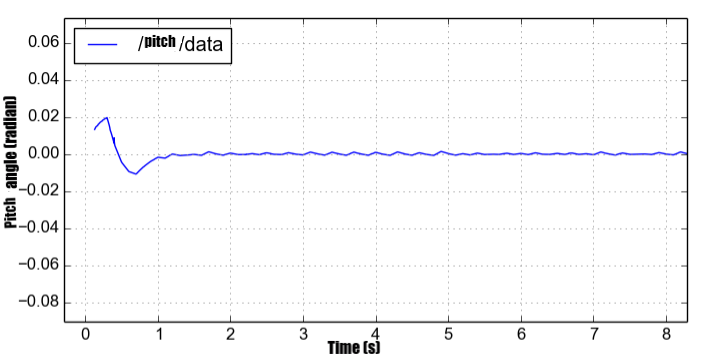}\\
\small (a) Kp=25, Ki =0.8, Kd =0.1
\end{tabular}
\begin{tabular}{c}
\includegraphics[width=0.5\textwidth]{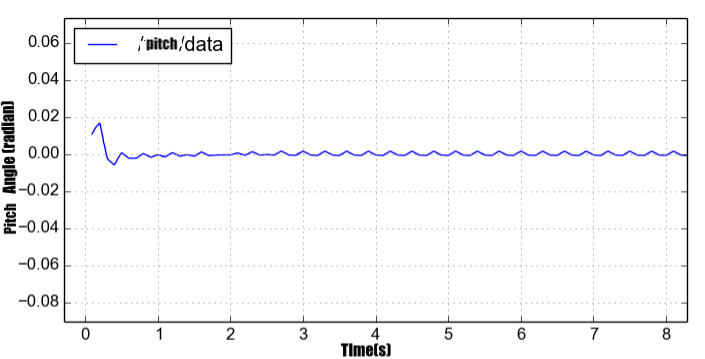}\\
\small (b) Kp=50 Ki=0.8 Kd =0.05
\end{tabular}
\end{figure}
\subsection{Fuzzy Logic}
\begin{figure}
\label{fig:FuzzyMember}
\begin{tabular}[b]{c}
\includegraphics[width=0.5\textwidth]{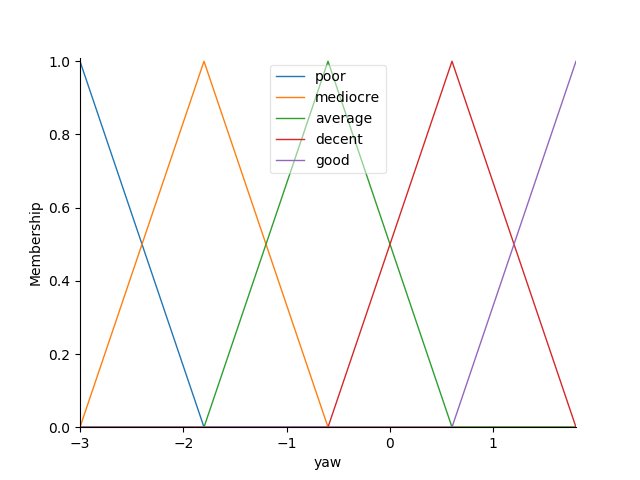}\\
\small (a) Pitch Angle of the robot
\end{tabular}
\begin{tabular}[b]{c}
\includegraphics[width=0.5\textwidth]{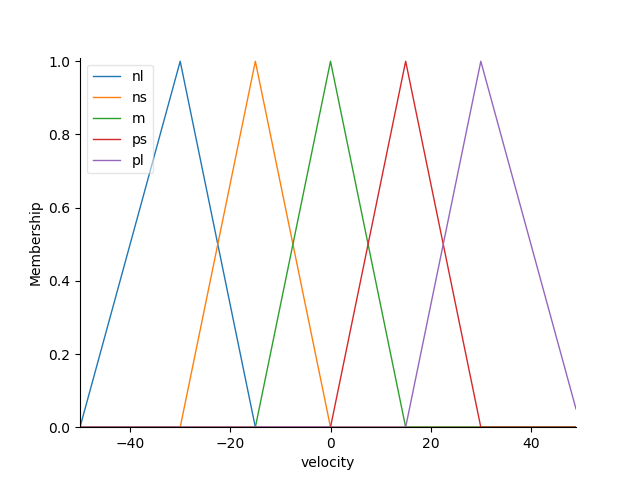}\\
\small (b) Pitch velocity of the robot
\end{tabular}
\caption{Fuzzy Membership Function}
\end{figure}

Fuzzy logic controllers are one of the most robust control systems in the world. In the academia , it was first described by Chuen Chien Lee \cite{man:Fuzzy}. There are many advantages of Fuzzy logic controllers. First of all, they can work on Multiple Input Multiple Output (MIMO) systems. Secondly, one doesn't need to derive the plant models for tuning the controllers. In this paper we have developed fuzzy PD and fuzzy PD+I controllers based on \cite{man:scikit-fuzzy}.  The rule base is given in Table~\ref{tab:FuzzyRule}. The membership functions for the velocity and yaw angle of the robot are given in Figure~\ref{fig:FuzzyMember}
\begin{table}[h]
\centering
\caption{Fuzzy Rule Base}
\label{tab:FuzzyRule}
\begin{tabular}{llllll}
\hline
$e'/e$         & Long Negative  & Short Negative & Medium & Short Positive & Long Positive  \\ 
Long Negative  & High Negative  & Negative & Zero & Positive & Positive \\ 
Short Negative & High Negative  & Negative  & Zero & High Positive  & High Positive  \\ 
Medium         & High Negative  & High Negative  & Zero & High Positive  & High Poitive   \\ 
Short Positive & High Negative  & Negative & Zero & Positive & High Positive  \\ 
Long Positive  & Negative & Negative & Zero &  Positive & High Positive  \\ 
\end{tabular}
\end{table}

\begin{figure}

\begin{tabular}[b]{c}
\includegraphics[width=0.5\textwidth]{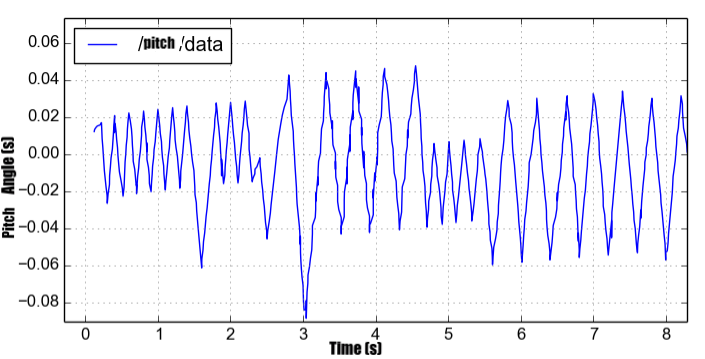}\\
\small (a) Fuzzy PD
\end{tabular}
\begin{tabular}[b]{c}
\includegraphics[width=0.5\textwidth]{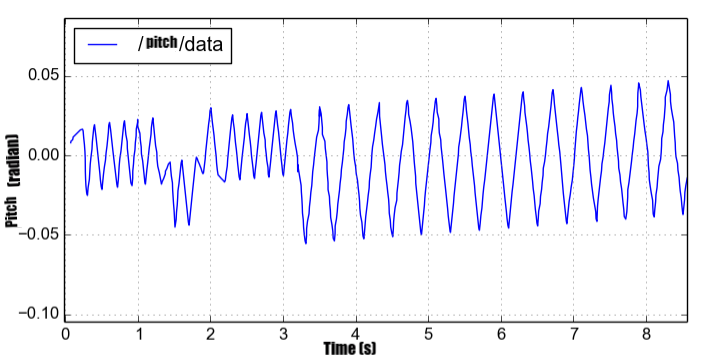}\\
\small (b) Fuzzy PD+I
\end{tabular}
\caption{Performance Curves for Fuzzy Logic Controller}
\label{fig:Fuzzy}
\end{figure}
\begin{figure}
\begin{tabular}[b]{c}
\includegraphics[width=0.5\textwidth]{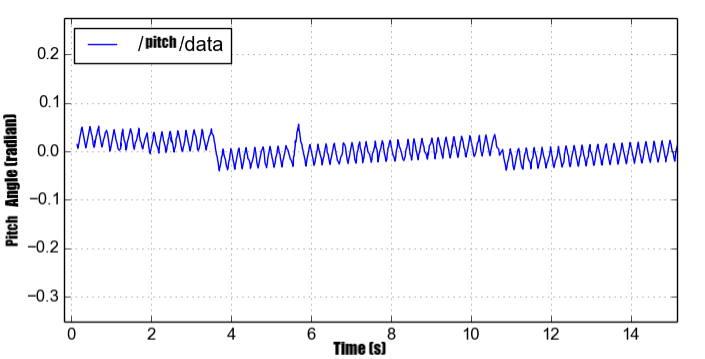}\\
\small (a)  $Q_1$ and $R_1$
\end{tabular}
\begin{tabular}[b]{c}
\includegraphics[width=0.5\textwidth]{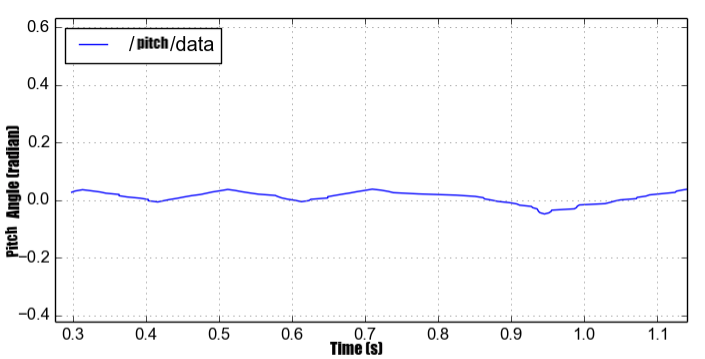}\\
\small (b) $Q_2$ and $R_2$
\end{tabular}
\caption{Performance Curves for  LQR Controller}
\label{fig:LQR}
\end{figure}
\subsection{LQR}
Another Robust Controller is Linear Quadratic Regulator Controller. H. Wang et al \cite{man:LQR} had discussed about the Implementation of LQR controllers on Inverted pendulum. Also \cite{man:michigan} has an extensive section about how to implement using Matlab. But , to the best of our knowledge , it has not been implemented in ROS simulation yet. The state space equations are given in \cite{man:michigan} as following, 
\begin{equation}\label{eq:stateSpace}
X = A+Bu
\end{equation}
Here, 
\begin{equation}
u = -KX
\end{equation}
, so the equation ~\ref{eq:stateSpace} becomes
\begin{equation}
X = (A-BK)X
\end{equation}
where ,
\begin{equation}
A= \left[\matrix{%
0 & 1 & 0 & 0 \cr
0 & \frac{-(I+ml^2)b}{I(M+m)+Mm^2} & \frac{m^2gl^2}{I(M+m)+Mm^2} & 0 \cr
0 & 0 & 0 & 1 \cr 
0 & \frac{-mlb}{I(M+m)+Mm^2} & \frac{mgl(M+m)}{I(M+m)+Mm^2} & 0
}\right]
\end{equation}
\begin{equation}
B = \left[ \matrix{0 \cr 
 \frac{I+ml^2}{I(M+m)+Mm^2} \cr 
 0 \cr 
 \frac{ml}{I(M+m)+Mm^2}
 } \right]
\end{equation}
\begin{equation}
[X = \left[ \matrix{\dot{x} \cr 
\ddot{x} \cr 
\dot{\phi} \cr 
\ddot{\phi} } \right]
\end{equation}
As we are more concerned with the yaw angles of the robot, not the linear movement and/or velocity , in our case the values of A,B and X will change. In our case , the values will become, 
\begin{equation}
A=\left[\matrix{ 0 & 1 \cr 
\frac{mgl(M+m)}{I(M+m)+Mm^2} & 0}\right]
\end{equation}
\begin{equation}
B = \left[ \matrix{  0 \cr 
 \frac{ml}{I(M+m)+Mm^2}  }\right]
 \end{equation}
\begin{equation}
X = \left[\matrix{\dot{\phi} \cr 
\ddot{\phi} }\right]
\end{equation}
From \cite{man:LQRCaltech} we get the cost function 
\begin{equation}
J = \int (x^TQx +u^TRu)dt
\end{equation}

Where , Q is the penalty matrix for the system, which penalizes the cost function for any error in the input. R penalizes the force input in the system. Now, the goal in the LQR controller is to choose $K$ matrix such that, we will have minimum cost function according to our need , i.e. according to our matrices $Q$ and $R$. In this project, we have experimented on two values of Q and R. For first one ,
\begin{equation}
 Q_1 = \left[\matrix{10 & 0 \cr 
0 & 100}\right]
\end{equation}
and 
\begin{equation}
R_1 = 0.001 
\end{equation}
The Second values are
\begin{equation}
Q_2= \left[\matrix{ 100 & 0 \cr 
0 & 1000}\right]
\end{equation}
and 
\begin{equation}
R_2 = 0.0001
\end{equation}
Figure~\ref{fig:LQR} shows the Response curves .
\section{COMPARISON TO PREVIOUS WORKS}
\subsection{Available ROS Controller}

In the available literature on how to use control systems with ROS , only PID control systems are available.They focus on how to use for different purpose, like Joint Controller, Position Controller, Effort Controller etc.\cite{man:ROSControl} But to the best of our knowledge no work exists on non PID controllers with ROS. Moreover, the existing work on PID Control system is not user friendly. Impossible to tune the controller gains while simulation is running.
The application of available ROS controller system is shown in Fig. ~\ref{fig:ROSControl} which is retrieved from  \cite{man:ROSControl}
\subsection{MATLAB and Simulink Simulation}

Matlab and Simulink are two of the most famous simulation softwares in Control systems and Robotics. But they have some problems
\begin{itemize}
\item To work on Matlab/Simulink simulation, one has to derive the control system of the robot before hand using physical laws and some assumptions. There is a high chance of error in this mathematical modeling , as it depends largely on the assumptions of those physical laws.
\cite{man:MATLAB}
In gazebo, there's no need to go through the mathematical model of the robot. The properties needed from systems like, mass, moment of inertia, center of gravity are used to build the robot model. The Physics Engine of Gazebo itself adjusts and derives the mathematical model from the robot model's properties.\cite{man:GazeboRef}

\item However, in Simmechanics one can import CAD models for simulation.\cite{man:Simmechanics} But it has it's own limits. 
\begin{itemize}
\item It is suitable for simple mechanical models.
\item It can only simulate mechanical properties , not the simulation for other sensors , like, camera, imu, laser sensors etc.
\end{itemize}
\item MATLAB and SIMULINK can only be used for control system simulation. But ROS and GAZEBO can be used like real world problems where simulation of control system, computer vision, Three dimensional positionin, path planning of the robots are needed.

\item The codes used in ROS simulation can directly be used in Real world robots, which is not the case for MATLAB and SIMULINK
\end{itemize}
 
\section{CONCLUSION}
The main purpose of this paper is develop explore and experiment different control theory on ROS simulation of Self Balancing Robot, which has been achieved successfully. The Gazebo model were built successfully. The real time data of the pitch angles vs time were plotted. From the Response curves, it can be noticed that , the most stable performance  was achieved using PID controller with the gains 50,0.8 and 0.05 . The faster response is the response curve of LQR controllers with First Q matrix and R value. The Fuzzy logic controller based performance is highly unstable. Which leads to our next research topic. Tuning the fuzzy rule base. In this project, we have written the rule base , based on intuition. Several algorithms exist in the literature to tune the correct rule base. Nevertheless, we hope that this project will be beneficial to the Robotics Researchers to learn more about the Control theory.

\end{document}